\documentclass{article}

\usepackage[utf8]{inputenc}
\usepackage{amsmath,amsfonts,amssymb}
\usepackage{graphicx}
\usepackage{mathtools} 
\usepackage{enumitem}  
\usepackage{authblk}
\usepackage[normalem]{ulem}  

\usepackage{amsthm}
\usepackage{algorithm}
\usepackage{algpseudocode}

\theoremstyle{definition}
\newtheorem{definition}{Definition}

\usepackage{xcolor}
\colorlet{lightgray}{gray!20}

\newcommand{\graybox}[1]{
  \colorbox{lightgray}{
    \par\noindent
    \begin{minipage}{\textwidth-1.5cm}#1\end{minipage}
  }\vskip\parskip}

\ifx \DRAFT \undefined 
\usepackage[disable]{todonotes}  
\else 
\usepackage{todonotes} 
\fi

\usepackage{xargs}    

\newcommand{\comment}[1]{\todo[linecolor=gray,backgroundcolor=gray!25,bordercolor=green]{\textbf{Comment:} #1}}
\newcommand{\actualized}[1]{\todo[linecolor=red,backgroundcolor=gray!25,bordercolor=red]{\textbf{Actualized:} #1}}

\newcommand{\passnum}[1]{$a_{p{#1}}$}
\newcounter{popass}
\newenvironment{popass}[1][]{
\refstepcounter{popass}\par
   \noindent{\boldmath\passnum{\thepopass#1}} \rmfamily {\medskip}}
   
\newcommand{\graypopass}[1]{\begin{popass}\graybox{#1}\end{popass}}

\newcommand{\sassnum}[1]{$a_{s{#1}}$}
\newcounter{spass} 
\newenvironment{spass}[1][]{\refstepcounter{spass}\par
   \noindent{\boldmath\sassnum{\thespass#1}}\rmfamily}  

\newcommand{\grayspass}[1]{\begin{spass}\graybox{#1}\end{spass}}

\newcommand{\assnum}[1]{$a_{#1}$}
\newcounter{ass}
\newenvironment{ass}[1][]{\refstepcounter{ass}\par
   \noindent{\boldmath\assnum{\theass#1}} \rmfamily}

\newcommand{\grayass}[1]{\begin{ass}\graybox{#1}\end{ass}}

\newcommand{\initassnum}[1]{$a^0_{#1}$}
\newcounter{initass}
\newenvironment{initass}[1][]{\refstepcounter{initass}\par
   \noindent{\boldmath\initassnum{\theinitass#1}} \rmfamily}

\newcommand{\grayinitass}[1]{\begin{initass}\graybox{#1}\end{initass}}

\newcommand{\setvalue}[2]{
    \ifdefined #1
        \renewcommand{#1}{#2}
    \else
        \newcommand{#1}{#2}
    \fi
}

\newcommand{\currentver}{1.3}
\newcommand{\version}{\currentver}

\title{
\ifx \DOCUMENTATION \undefined 
Formal specification terminology for  \newline 
demographic agent-based models  of \newline 
fixed-step single-clocked simulations  
\newline \newline 
\else 
Specification of MiniDemographicABM.jl: \\ A simplified agent-based demographic  model \\  
of the United Kingdom \newline  
 { \normalsize (Version \version) } 
\fi 
\author{Atiyah Elsheikh}
\affil{MRC/CSO Social and Public Health Sciences Unit, University of Glasgow, Glasgow, United Kingdom \\ \vspace{0.5cm}
atiyah.elsheikh@glasgow.ac.uk}
\date{\today}
}

\begin{document}

\maketitle

\noindent 
\ifx \undefined \DOCUMENTATION
\begin{abstract}

\noindent 
This document presents adequate formal terminology for the mathematical specification of a subset of Agent Based Models (ABMs) in the field of Demography. 
The simulation of the targeted ABMs follows a fixed-step single-clocked pattern.
The proposed terminology further improves the model
understanding and can act as a stand-alone protocol for the specification and optionally the documentation of a significant set of (demographic) ABMs.  
Nevertheless, it is imaginable the this terminology can serve as an inspiring basis for further improvement to the largely-informal widely-used model documentation and communication O.D.D.\ protocol \cite{Grimm2020,Amouroux2010} to reduce many sources of ambiguity which hinder model replications by other modelers.
A published demographic model documentation, largely simplified version of the Lone Parent Model \cite{Gostoli2020} is separately published in \cite{Elsheikh2023} as illustration for the formal terminology presented here. The model was implemented in the Julia language \cite{MiniDemographicABMjlMisc} based on the Agents.jl julia package \cite{Datseris2022}. 

\end{abstract}
\else 
\begin{abstract}

\noindent 
This documentation specifies a simplified non-calibrated demographic agent-based model of the UK, a largely simplified version of the Lone Parent Model presented in \cite{Gostoli2020}. 
In the presented model, individuals of an initial population are subject to ageing, deaths, births, divorces and marriages throughout a simplified map of towns of the UK.  
The specification employs the formal terminology presented in \cite{Elsheikh2023a}. 
The main purpose of the model is to explore and exploit capabilities of the state-of-the-art Agents.jl Julia package \cite{Datseris2022} in the context of demographic modeling applications. 
Implementation is provided via the Julia package MiniDemographicABM.jl \cite{MiniDemographicABMjlMisc}.
A specific simulation is progressed with a user-defined simulation fixed step size on a hourly, daily, weekly, monthly basis or even an arbitrary user-defined clock rate. 
The model can serve for comparative studies if implemented in other agent-based modelling frameworks and programming languages.
Moreover, the model serves as a base implementation to be adjusted to realistic large-scale socio-economics, pandemics or immigration studies mainly within a demographic context. 
\end{abstract} 
\fi 

\ifx \undefined \DRAFT 
\else 
\ifx \undefined \DRAFT
\else
\section*{Remarks}

The side notes (s.a.\ this\actualized{Version 1.0 this is used to show recent updates and changes)} can be disabled by commenting the macro: 
\begin{verbatim}
\def \DRAFT {} 
\end{verbatim} at the top of the tex document. \\ 
\fi

\ifx \undefined \DOCUMENTATION
List of todo-s for paper 
\begin{itemize}
\item fix assumption numbers 
\item Selection of journal \comment{This is not a software paper but still it is worth to have a look at \url{https://www.software.ac.uk/which-journals-should-i-publish-my-software}}
\item abstract 
\item fix motivation and contribution highlight 
\item \sout{potential case studies}  
\item improve outlook 
\item Related works , e.g.\ related to temporal logic and verification of hybrid systems 
\item Reduce the paper contents by citing the model documentation and emphasizing the highlights 
\begin{itemize}
    \item Make an additional appendix for O.D.D?
\end{itemize}
\item Examine if it is possible to follow O.D.D-like guidelines (In an Appendix)
\item Citations 
\item Outlook: extension to this work to other type of multi agents based model (multi-level, multi clocks, dynamic scheduling etc.) 
\end{itemize}

Suggested citations: 
\begin{itemize}
    \item 
    Open science is a research accelerator., Nature chemistry 3 (2011)
     \item 
     Working Practices, P.\ v B, Scientists and software engineers: a tale of two cultures, 2019
     \item 
     Wilkinson, M.D., et al.: The FAIR guiding principles for scientific data management and stewardship. Sci. Data 3(1), 160018 (2016)
     \item 
     The ODD Protocol for Describing Agent-Based and Other Simulation Models: A Second Update to Improve Clarity, Replication, and Structural Realism, 
     \item 
     The ODD Protocol for Describing Agent-Based and Other Simulation Models: A Second Update to Improve Clarity, Replication, and Structural Realism' Journal of Artificial Societies and Social Simulation 23 (2) 7 <http://jasss.soc.surrey.ac.uk/23/2/7.html>. doi: 10.18564/jasss.4259
     \item 
     O.D.D.: a Promising but Incomplete Formalism For Individual-Based Model Specification
\end{itemize}
\else 
\begin{itemize}
    \item use of algorithms 
    \item Appendix as O.D.D. 
\end{itemize}
\fi 
\fi 

\ifx \undefined \DOCUMENTATION
\section{Terminology}
\label{sec:terminology}

This section introduces the basic terminological foundation on which fundamental terminology is established for the specification of ABMs and their simulation process following fixed-step single-clocked scheduling.  

\subsection{Populations $P = M \cup F$}
\label{sec:terminology:population}

Given that $F(t) \equiv F^t  \equiv F~(M(t) \equiv M^t$) is the set of all females (males) in a given population $P(t) \equiv P^t$ at an arbitrary time point $t$ where 
\begin{equation}
\label{eq:population}
    P(t) = M(t) \cup F(t)
\end{equation}
$t$ is sometimes omitted for the sake of simplification i.e.\ $P \equiv P^t$. 
In this case, $P$ refers to the set of all individuals at an implicitly known time point $t$ (e.g.\ the current simulation iteration). \\ 

Analogously, 
$t$ is sometimes placed as a superscript (e.g.\ $F^t$) purely for readability purpose, e.g.\ in the context of algorithm specification. 
Similarly, an individual $p \in P(t)$ when attributed with time, i.e.\ $p^t$, refers to that individual at time point $t$.  \\ 

Furthermore, 
explicit specification of a time interval is realizable as follows: 
\begin{equation}
\label{eq:population:interval} 
P([t_1,t_2]) \equiv P^{[t_1,t_2]}
\end{equation}
indicating the set of all population individuals between $t_1$ till $t_2$ inclusive.
If the interval $[t_1,t_2]$ is associated with a step-size, i.e.\ $[t_1,t_2]_{\Delta t}$, then this time interval is discretized in equidistant time points of length $\Delta t$ starting from $t_1$ till the last possible time point.

\subsection{Population features $\mathcal{F}$}
\label{sec:features}

Every individual $p \in P$ is attributed by a set of features $f \in F$.    
Examples of elementary population features can span but not limited to the following: 
\begin{itemize}
\item \underline{age}, e.g.\ particular age group e.g.\ neonates, children, teenagers, thirties, etc. or specific age e.g.\ 25 years old 
\item \underline{alive status}, i.e. whether alive or dead
\item \underline{gender}, e.g.\ male or female 
\item \underline{space}, e.g.\ inhabitant of a particular town 
\item \underline{kinship status} or relationship, e.g. father-ship, parents, orphans, divorcee, singles etc. 
\end{itemize}

\subsection{Features expressed as predicates}
\label{sec:terminology:predicates}

Population or a person features are explicitly expressed in terms of predicates of various types depending on the outcomes s.a.\ 
\begin{enumerate}
    \item Boolean predicates s.a. $isMale?$, $age>45?$ or $livesInGlasgow?$ 
    \item Individual predicates s.a.\ $fatherOf$  
    \item Grouping predicates s.a.\ $siblingsOf$ or $childrenOf$
\end{enumerate}
The naming choice of the predicate mostly indicates the type of the predicate. However, while nothing prevents the syntactical exchange the predicate "$fatherOf$" with "$father$", the context may lead to an ambiguity in case "$father$" is interpreted as "$isFather?$". \\ 

In this case, it is recommended that the type of the predicate to be explicitly specified by exploiting the post-fix "?" s.a. "$father?$". Similarly, there is nothing against removing the postfix "?" in the predicate $hasSiblings?$ since it is obviously excessive. \\

Nevertheless, the confusion may go further due to the interpretation of $hasSiblings$ whether it implies has more than one sibling or any sibling. Here if the context is not clear, then further precisions need to be employed s.a.\ $hasASibling$ and $hasMoreThanOneSibling$.  \\ 

Formally, a given feature 
\begin{equation}
\label{eq:featureDef}
    f : P \mapsto X
\end{equation}
intuitively maps a given individual $p \in P$ to an element or a subgroup in another subset $X$. The values of the set $X$ depends on the predicate type correspondingly as follows:
\begin{enumerate}
    \item $\{ true, false \}$ for Boolean predicates: e.g.\ $isMale(Jonas \in P) = true$
    \item $X \subset P$ with $|X| = 1$ for individual predicates, e.g.\  $mother(Jonas) = \{Jasmine\} $ ($|*|$ stands for the size)
    \item $X \subseteq P$ for group predicates, e.g.\ $parents(Jonas) = \{ Jahua, Johanna \} $     
\end{enumerate}
The case that Jonas has no siblings is formulated as 
$$siblings(Jonas) = \phi$$ 
with $\phi$ standing for the empty group.

\subsection{Featured sub-populations via Boolean predicates $P_f$ } 
\label{sec:terminologies:fsubpopulations}

Boolean predicates are further exploited for the specification of featured sub-populations. 
Namely, let $P_{f}$ correspond to the set of all individuals who satisfy a given feature $f$. That is, if 
\begin{equation}
\label{eq:fp}
f(p \in P) = b \in \{true,false\}    
\end{equation}
Then
\begin{equation}
\label{eq:Pf}
P_f = \{ p \in P \text{ s.t.\ } f(p) = true \}    
\end{equation}

\subsubsection*{Examples}
\begin{itemize}
    \item $M = P_{male}$ \footnote{it is pre-known that "$male$" indicates a Boolean predicate and thus no need to employ "$isMale?$"} 
    \item $W_{married}$ 
    corresponds to the set of all married women
    \item $P_{age \geq 65}$ corresponds to all individuals of age older than 65
\end{itemize} 

\graybox{
\begin{definition}[Closed set of features]
\label{def:closedSetOfFeatures}
For the given set of features specified in Section \ref{sec:features}, a subset of features
\begin{equation}
\label{eq:closedSubsetElemFeatures}
F' = \{f'_1,f'_2,\dots \,f'_m\} ~\subset F    
\end{equation}
is called \emph{a closed subset of elementary features}, if 
the overall population constitutes of the union of the underlying elementary featured sub-populations, i.e.\ 
\begin{align}
\label{eq:PUnionPf}
P = P_{\mathcal{F'}} \equiv P_{f'_1 \cup f'_2 \cup \dots \cup f'_m} = P_{f'_1} \cup P_{f'_2} \cup \dots \cup P_{f'_m}     
\end{align} 
where $P_{f'_i} \cap P_{f'_j} = \phi$ for any valid subscripts $i,j$ and $i \neq j$.
\end{definition}
}
For example, male and female gender features constitute a closed set of elementary features as indicated by Equation \eqref{eq:population}.

\subsection{Featured sub-populations using group predicates $g(P)$}
\label{sec:terminology:gsubpopulations}

Analogously, as implied by Equation \ref{eq:featureDef}, group predicates are employed for specification of sub-populations: 
\begin{equation}
    \label{eq:gP}
    g(P) = \{ p \in P \text{ s.t.\ } \exists~ q \in P \text{ with } g(q) = p \} 
\end{equation}

\subsubsection*{Examples}

\begin{itemize}
\item 
    $children(\{Jad, Jasmine\})$ correspond to children of Jad and Jasmine 
\item 
    $mother(P)$
describes the set of all mothers of a given population 
\end{itemize}
Note that the last example corresponds to individual predicate which is indeed a special case of group predicates. The last example can be equivalently  expressed in terms of Boolean predicates:
$$
mother(P) \equiv P_{isMother} \equiv P_{mother}
$$
This may sound excessive, but the following is a example demonstrating the descriptive power when combining Boolean and group (or individual) predicates together for enabling a concise specification: 
$$
mother(P_{age \leq 3}) \equiv P_{motherWithChildrenOfAgeLessThanOrEqualToThree}
$$

\section{Composite features $\bigcup \mathcal{F}$}
\label{sec:compositeFeatures}

\subsection{Non-elementary features $f_i ~o~ f_j$}
\label{sec:terminologies:nonelemfeatures}

For a given set of elementary features $\mathcal{F}$ (informally an elementary feature demands only one descriptive predicate), the set of all non-elementary features $\bigcup \mathcal{F}$ is defined as follows:

\graybox{
\begin{definition}
A non-elementary feature:
$$f^* \in \bigcup \mathcal{F} 
~\text{ where }~ \mathcal{F} \subset \bigcup \mathcal{F}
$$ 
is recursively established from a finite number of arbitrary elementary features $$f_i,f_j,f_k, ... \in \mathcal{F}$$ by (but not limited to)
\begin{itemize}
    \item union (e.g.\ $f_i \cup f_j$  ),
    \item intersection (e.g.\ $f_i \cap f_j$)
    \item negation (e.g. $ \neg f_i$)
    \item exclusion or difference (e.g.\ $f_i - f_j$ ) 
\end{itemize}
\end{definition}
}
Formally, if 
$$f^* = f_i ~ o ~ f_j ~~~\text{ where }~~~ o \in \{ \cup, \cap , -\} \text{ and } f_i,f_j \in \bigcup \mathcal{F} $$
then 
\begin{equation}
\label{eq:Pfog}
P_{f^*} = P_{f_i~o~f_j} = \{ p \in P ~ s.t.\ p \in (P_{f_i} ~o~ P_{f_j}) \}  
\end{equation}
Analogously,
\begin{equation}
\label{eq:neg}
P_{\neg f}  = \{ p \in P ~ s.t.\ p \notin P_f   \} ~\text{ with }  f \in  \bigcup \mathcal{F}     
\end{equation}

Generally, any of the features $f_i$ and $f_j$ in Equation \eqref{eq:Pfog} or $f$ in \eqref{eq:neg} can be either elementary or non-elementary and the definition is recursive allowing the construction of an arbitrary set of non-elementary features. 

\subsubsection*{Examples}

Negation operator s.a.\ ($\neg$) is beneficial for sub-population specification, e.g. 
\begin{equation}
\label{eq:intersectionNegationEx}
F_{married ~\cap~ \neg hasChildren}     
\end{equation}
corresponds to all married females without children. This sub-population can be equivalently described using the difference operator: 
\begin{equation}
\label{eq:differenceEx}
F_{married ~-~ hasChildren}    
\end{equation}
which entails to be a matter of style unless algorithmic execution details of the operators are assumed\footnote{e.g.\ if assumed that within an intermediate computation the ($-$) operator is executed directly on the set of married females rather than the set of all females as the case when employing ($\cap$) instead}. \\

As another example, the sub-population  
\begin{equation}
\label{eq:example:nonelementary}
M_{divorced ~ \cap ~ hasChildren ~ \cap  ~ age>45 ~ - ~ hasSiblings}
\end{equation}
corresponds to the set of all divorced men of age older than 45 who has no siblings but they have children. 
Equation \eqref{eq:example:nonelementary} can be re-written as:
\begin{equation}
\label{eq:example:nonelem:readible}
M_{divorced} ~ \cap ~ M_{hasChildren} ~ \cap ~ M_{age>45} ~ - ~ M_{hasSiblings}     
\end{equation}
Both styles can be mixed together purely for readability or formatting purposes. 

\subsection{Composition of Boolean predicates $P_{f_1(f_2)}$} 
\label{sec:compositionBooleanOperators}

Another beneficial operator is the composition operator analogously defined as 
\begin{equation}
\label{eq:Pfg}
P_{f_1(f_2)} = \{ p \in P_{f_1} \text{ s.t.\ } f_2(p) = true \}
\end{equation}
where both $f_1$ and $f_2$ correspond to Boolean predicates. 
The composition operator can be regarded to be more  computationally efficient in comparison with the intersection operator\footnote{In this work, the main purpose behind the composition operator mainly remains in the context of algorithmic specification rather than enforcing any implementation details regarding computational efficiency}. \\

The desired sub-population specification in the example given by Equation \ref{eq:example:nonelem:readible} may not correspond to the desired specification. Namely, desired is to specify the alive divorced male population older than 45 years with alive children and alive siblings. 
In this case, the employment of the composition operator is relevant:
\begin{equation}
\label{eq:whatelse}
 M_{alive(divorced ~\cap~ hasAliveChildren ~\cap~ age>45 ~-~ hasAliveSibling)}   
\end{equation}

\subsection{Composition of group and Boolean $g(P_f)$ or ${g(P)}_f$}
\label{sec:compositionGroupBooleanOperators}

Alternatively, if $g$ corresponds to a group predicate , and $f$ corresponds to a Boolean predicate, then composition expresses the following:  
\begin{align}
\label{eq:fPg}
g(P_f) ~=~ & \{ p  \in P \text{ s.t.\ } \exists q \in P_f~ (\text{i.e.\ } f(q) = true) \text{ with } p \in g(q)\}  
\end{align}
For example, $children(M_{alive})$ corresponds to the children of the alive male populations (or equivalently population with alive fathers). Similarly, 
\begin{align}
\label{eq:fPUg}
g(P)_{f} ~=~ & \{ p \in g(P) \text{ s.t.\ } f(p) = true \} \\
& \text{ where } g(P) = \{ p  \in P \text{ s.t.\ } \exists q \in P \text{ with } p \in g(q) \} \nonumber 
\end{align}
Here, $children(M)_{alive}$ corresponds to the alive children of the male population, that is the composition operator should be interpreted with care as
\begin{equation}
\label{eq:non-symmetry}
children(M_{alive})  ~\not \equiv ~ children(M)_{alive}     
\end{equation}
The alive population with alive fathers is expressed as $children\left(M_{alive}\right)_{alive}$

\subsection{Composition of group operators $g_1(g_2(P))$}
\label{sec:compositionGroupOperators}

As a matter of completeness and as an illustrative example, the following phrase expresses the grandchildren of all population of age larger than 65
\begin{equation}
\label{eq:groupOperators}
    children(children(P_{age>65})) \equiv grandChildren(P_{age>65}) 
\end{equation}
However, nothing is against employing the predicate $grandChildren$ directly. 

\todo{Section special cases for $borther(p_{alive})$ or  }

\section{Temporal operators}
\label{sec:temporal}

This section introduces further operators, inspired by the field of temporal logic. 
These operators provide powerful capabilities for algorithmic specification of complex phrases with temporal elements in a compact manner. 
The demonstrated temporal operators are included in the set of composite features $\bigcup \mathcal{F}$. 

\subsection{just operator}
\label{sec:temporal:just}

A special operator is 
\begin{equation*}
just(P_f) \equiv P_{just(f)} \subseteq P_f ~,~ f \in \bigcup \mathcal{F} 
\end{equation*}
standing for a featured sub-population established by an event that has just occurred (in the current simulation iteration).
For instance,
$$
just(P_{married}) \equiv 
P_{just(married)}^{t + \Delta t}  ~~~\text{ where } ~~~ \Delta t: \text{ a simulation fixed step size } 
$$ 
stands for those individuals who just got married in the current simulation iteration but they were not married in the previous iteration, i.e.
$$
P_{just(married)}^{t+\Delta t} ~ = ~  P_{married}^{t+\Delta t} ~ - ~ P_{married}^t 
$$
Formally, 
\begin{equation}
\label{eq:just}
P_{just(f)}^{t + \Delta t} = 
P_{f}^{t+\Delta t} ~ - ~ P_{f}^t     
\end{equation}
The just operator provides powerful capabilities for concise specification when combined with the negation operator. 
For example, 
$$P_{just(\neg married)}^{t+\Delta t}$$ 
stands for those who "just" got divorced or widowed.

\subsection{pre operator}
\label{sec:temporal:pre}

Another distinguishable operator is 
$$pre(P_f) \equiv P_{pre(f)} ~,~ f \in \bigcup \mathcal{F}$$ 
standing for a featured $f-$sub-population in the "previous" iteration. 
So for instance,
$$P_{pre(married)}^{t+\Delta t}$$ 
stands for those individuals who were married (and not necessarily just got married) in the previous simulation iteration
$$
pre(P_{married}^{t + \Delta t}) ~ \equiv ~ 
P_{pre(married)}^{t+\Delta t} ~ \equiv ~ P_{married}^t 
$$
Formally, 
\begin{equation}
\label{eq:pre}
   P_{pre(f)}^{t + \Delta t} =  P_f^t  
\end{equation}
Temporal operators can also be applied to individuals and their attributes. For instance 
$$pre(location(p \in P^t)) = Glasgow $$ 
stands for the location of a person in the previous iteration (which does not need to be either similar or different in the current iteration). \\ 

This operator may look unnecessary excessive, however the demographic model specification \cite{Elsheikh2023} of the package implemented in \cite{MiniDemographicABMjlMisc} makes use of the $pre$ operator several times. For instance, a model assumption related to a divorce event is informally and formally described as follows: 
\begin{center}
\graybox{
Any male who just got divorced moves to an empty house within the same town:
\begin{align}
 m \in & M_{just(isDivorced}^t) \implies \nonumber \\  
& pre(location(m)) \neq location(m) ~\text{ and }~ livesAlone(m) ~\text{ and }~ \nonumber \\  
& town(m) = pre(town(m))  
\end{align}
}
\end{center}


\section{post operator} 

The operator $post$ is similar to the $pre$ operator, but it is rather concerned with the "post" iteration. 
Without loss of information, the formal description is analgous to the $pre$ operator.  
The following is an example from the accompained demonstrated model for formulating a model assumption making use of the $post$ operator: 

\begin{center}
\graybox{
Once a new house is built (e.g.\ to host an adult moving out of his parent's house), it never gets demolished and remains always inhabitable 
\begin{equation}
\label{eq:ass:HouseAlywasRemain}
    \text{ if } h \in H(t) ~\implies h \in post(H(t)) 
    \text{ where } t_0 \leq t < t_{final}
\end{equation}
}
\end{center}
\section{General form}
\label{sec:generalform}

\subsection{General definitions}
\label{sec:general:definitions}

This article is concerned with proposing a terminology for the mathematical specification of demographic ABMs based on a single-clocked fixed-step simulation formally defined via the tuple 
\begin{equation}
  \label{eq:abmDefinition}<\alpha_{sim},\mathcal{F},\mathcal{M},\mathcal{M}^{t_0},\mathcal{E},\mathcal{A},\mathcal{A}^{t_0}>  
\end{equation}
where 
\begin{itemize}
\item 
$\alpha_{sim} = (\Delta t, t_0, t_{final}, \alpha_{meta})^T$:
simulation parameters including a fixed step size and final time-step after which the simulation process ends  
\begin{itemize}
\item $\alpha_{meta}$:
Implementation-dependent simulation parameters, e.g.\ simulation seed for random number generation 
\end{itemize}
\item 
$\mathcal{F}$: a finite set of elementary population features, cf.\ Section \ref{sec:generalform:features}
\item 
$\mathcal{M}$:  a mathematical model representation corresponding to (demographic) ABM, cf. Section \ref{sec:generalform:abm}  
\item 
$\mathcal{M}^{t_0}$: a mathematical model that evaluates the initial model state at time $t_0$, cf.\ Section \ref{sec:generalform:initialstate}  
\item 
$\mathcal{E}$: a finite set of events that takes place in the population. They transient the states of the (featured sub-)population(s), cf.\ Section \ref{sec:generalform:events}
\end{itemize}
Establishing a mathematical model that corresponds to reality till the tiniest details is impossible. Therefore, a set of (typically non-realistic) assumptions has to be included in order to simplify the model specification process:
\begin{itemize}
    \item 
$\mathcal{A}$: a set of model assumptions that should not be violated during the simulation course between $t_0$ and $t_{final}$, cf.\ Section \ref{sec:generalform:Assumptions}
\item 
$\mathcal{A}^{t_0}$: a set of initial model assumptions that should not violate the initial model state $\mathcal{M}^{t_0}$, cf.\ Section \ref{sec:generalform:initialAssumptions} 
\end{itemize}

\subsection{Population features $~\mathcal{F}~$} 
\label{sec:generalform:features}

\underline{$\mathcal{F}= \{f_1,f_2,f_3,...f_k\}$} describes a finite set of elementary features each distinguishes a featured sub-population $$P_f(t) \subseteq P(t)  ~ , ~ f \in \mathcal{F}$$
as defined in Equations \ref{eq:fp} and \ref{eq:Pf}, cf.\ Section \ref{sec:terminologies:fsubpopulations} for examples of population features.

\subsection{Demographic agent-based model $\mathcal{M}$}
\label{sec:generalform:abm}

\underline{$\mathcal{M}$} corresponds to a demographic ABM formally defined as:  
\begin{align}
\label{eq:abmModel}
& \mathcal{M} \equiv \mathcal{M}^t \equiv \mathcal{M}(P,S,\alpha,D,t)   \\
\label{eq:abmModelInitialConditions}
& \text{ associated with } \mathcal{M}(P(t_0),S(t_0),\alpha,D(t_0),t_0) =  M^{t_0} 
\end{align}
where 
\begin{itemize} 
\item \underline{$P \equiv P(t)$}: a given population of agents (i.e.\ individuals) at time $t$ evaluated via the model $\mathcal{M}(t)$
\item 
\underline{$S \equiv S(t)$}: 
the space on which individuals $p \in P$ are operating  
\item 
\underline{$\alpha$}: 
time-independent model parameters 
\item 
\underline{$D(t)$}: 
input data integrated into the model as (possibly smoothed) input trajectories 
\end{itemize} 
In \cite{MiniDemographicABMjlMisc}, the space is set as: 
\begin{equation}
\label{eq:spaceExample}
S(t) = <H(t), W> 
\end{equation}
where $H(t)$ stands for a set of houses distributed within the the set of towns $W$.

\subsubsection*{Featured sub-populations (via $ \mathcal{M}_{f^*} ~,~ f^* \in \bigcup \mathcal{F}~$)} 

This subsection concerned with featured sub-populations used to distinguish sub-populations needed for specification of the transient processes via events $\mathcal{E}$ within the agent-based modeling simulation process. 
A featured sub-population $P_f, f \in \mathcal{F}$ is evaluated by the sub-model $M_{f}$ concerned only with the elementary features $f$:
\begin{equation}
\label{eq:Mf}
f(\mathcal{M}) \equiv f(\mathcal{M}^t) \equiv \mathcal{M}_f^t = \mathcal{M}_{f}(P_{f},S,\alpha,D,t)     
\end{equation}
evaluating or predicting the sub-population 
\begin{equation}
\label{eq:generalform:fP}
f(P(t)) \equiv P_f(t) ~\text{ s.t.\ }~ \forall p \in P_f(t) \implies f(p) = true     
\end{equation}
For a given closed set of elementary features as given in Equation \ref{eq:closedSubsetElemFeatures}, the overall population is the union of these elementary features, 
cf.\ Equation \ref{eq:PUnionPf}. 
In that case, the comprehensive model $\mathcal{M}$ constitutes of the sum of its elementary featured sub-models:
\begin{equation}
\label{eq:MSumMf}
\mathcal{M} \equiv \sum_{f' \in \mathcal{F'}} \mathcal{M}_{f'}  
\end{equation}
Note that this terminology extends to composite features $\bigcup \mathcal{F}$ as well by rather assuming $f \in \mathcal{\bigcup F}$ in Equation \eqref{eq:Mf}.

\subsection{Initial population and space (via $\mathcal{M}^{t_0}$)}
\label{sec:generalform:initialstate}

\underline{$\mathcal{M}^{t_0}$} is a model that evaluates the initial population $P(t_0)$ and the initial space $S(t_0)$ at a proposed simulation start time $t_0$ via Equation \eqref{eq:abmModelInitialConditions}. 
Consequently, the state of the model, i.e.\ $P(t)$ and $S(t)$, depends on the initial conditions expressed as $M^{t_0}$.  \\ 

Additionally, the initial state of the model $M^{t_0}$ combined with the elementary population features evaluate elementary as well as composite featured sub-populations:  
\begin{equation}
\label{eq:Mft0}
f(M^{t_0}) = M_f^{t_0}, \forall f \in \bigcup \mathcal{F}    
\end{equation}
Consequently, both the corresponding initial population and featured sub-populations:
\begin{equation}
\label{eq:Pt0Pft0}
P(t_0) \text{ and } P_f(t_0), ~~~ \forall f \in \mathcal{F}
\end{equation}
 are specified, e.g.\ the initial ratio of females and males, the age distribution of the population, the spatial distribution of the population, among others. \\ 

In the model described by the package implementing the documentation given in \cite{MiniDemographicABMjlMisc}, the initial space is set as  
\begin{equation}
\label{eq:St0}
    S(t_0) ~=~ <H(t_0)~,~W> 
\end{equation}
composed of a pair of an initial set of houses $H(t_0)$ distributed within a set of towns $W$, in conformance with the space setting in Equation \eqref{eq:spaceExample}.

\subsection{Events $\mathcal{E}$}
\label{sec:generalform:events}

\underline{$\mathcal{E} = \{e_1, e_2, e_3, ..., e_n\} $} is a finite set of events, each of which transients a particular featured sub-population $P_{f^*}(t)$ is evaluated by the sub-model:  
\begin{equation}
\label{eq:fstarM}
f_1^*(\mathcal{M}^t) \equiv \mathcal{M}_{f_1^*}(P_{f_1^*},S,\alpha,D,t) \text{ with } f_1^* \in \bigcup \mathcal{F} 
\end{equation}
to another modified sub-population predicted by 
\begin{equation}
f_2^*(M^{t+\Delta t}) \equiv \mathcal{M}_{f_2^*}(P_{f_2^*},S,\alpha,D,t + \Delta t)    
\end{equation}
Formally,
\begin{equation}
\label{eq:events}
e(\mathcal{M}_{f_1^*}^{t}) = \mathcal{M}_{f_2^*}^{t + \Delta t} ~~~\text{ for some } \{f_1^* , f_2^*\} \in \bigcup \mathcal{F} ~~\text{ where } e \in \mathcal{E}    
\end{equation}
The application of all events transients the model to the next state: 
\begin{equation}
\label{eq:stepping}
\prod_{i=1}^n e_i(\mathcal{M}^t) = \mathcal{M}^{t + \Delta t} 
\end{equation}

As an example, the model specification provided in \cite{Elsheikh2023a} specifies the set of events as
$$
\mathcal{E} = \{ ageing, birth, death, divorce, marriage \} 
$$
The death event specification transforms a given population of alive individuals to the same population with some individuals possibly dead: 
\begin{align}
\label{eq:event:deaths}
    death \left( P_{isAlive }^t \right)  ~=~  P_{isAlive - age=0}^{t+\Delta t}  ~\bigcup~  P_{just(\neg isAlive)}^{t+\Delta t} 
\end{align}
The first phrase in the right hand side stands for the alive population except neonates and the second stands for those who just became dead.

\subsection{Assumptions $\mathcal{A}(\mathcal{M}^{[t_0,t_{final}]})$}
\label{sec:generalform:Assumptions}

\underline{$\mathcal{A} = \{a_1, a_2, a_3, \dots, a_s\}$} is a finite set of assumptions each represented as a Boolean condition  
\begin{equation}
\label{eq:assumptions}
    a_i(\mathcal{M}^t) = 
    a_i(\mathcal{M}(P(t),S(t),\alpha,D(t),t)) = b \in \{true,false\}  
\end{equation}
if 
$a_i(M^t) = true$ one says that assumption $a_i$ satisfies $M^t$ expressed as follows: 
\begin{equation}
\label{eq:assumptionSatisfy}
a_i \vdash \mathcal{M}^t  ~~~,~~~ \forall i \in \{ 1,2,\dots,s \} 
\end{equation}
A particular assumption $a_j$ violates $\mathcal{M}$ if at any simulation time point 
$$t \in [t_0,t_{final}]_{\Delta t} \implies a_j(M^t) = false$$ 
expressed as: 
\begin{equation}
\label{eq:assumptionViolation}
a_j \nvdash \mathcal{M}^{[t_0,t_{final}]_{\Delta t}}   ~~~,~~~ \text{ where } j \in \{1,2,\dots,s\}    
\end{equation}
That is, the violation of an assumption may also depend on the simulation parameter resolution, i.e.\ $\Delta t$. 
Typically, an assumption would be usually rather concerned with a featured sub-population, i.e.\ 
\begin{equation}
\label{eq:assumptionRelatedToSubpopulation}
    a_i(\mathcal{M}^t_{f^*}) = 
    a_i(\mathcal{M}(P_{f^*}(t),S(t),\alpha,D(t),t)) = b \in \{true,false\}  ~,~f^* \in \bigcup \mathcal{F}
\end{equation}
A given assumption could be purely related to the space, the population (a sub-population), or both the space and the (or a sub-)population.  
Examples of assumptions are listed as follows: 

\begin{center}
\graybox{(Space-related assumption) 
Once a new house is built (e.g.\ to host an adult moving out of his parent's house), it never gets demolished and remains always inhabitable 
\begin{equation}
\label{eq:ass:HouseAlywasRemain}
    \text{ if } h \in H(t) \text{ and } t < t'  ~\implies h \in H(t')
\end{equation}
}
\graybox{
(Population-related assumption)
Only a married female\footnote{This was assumed in the lone parent model and obviously the marriage / partnership concept needs to be re-defined in the context of realistic studies} under age of 45 gives birth
\begin{align}
\label{eq:ass:pop:onlyMarriedGivesBirth}
f \in & F_{just(gaveBirth)}^t \implies \nonumber \\ & 
isMarried(f^t) = true \text{ and } age(f^t) < 45
\end{align}
}
\graybox{(mixed-assumption) 
A neonate's house is his mother house:
\begin{align}
p \in~ &  P_{age=0}   
\implies \nonumber \\ 
& house(p) = house(mother(p))    
\end{align}
}  
\end{center}
In the accompanied model specification, assumptions are numbered and labeled according to their types, namely,  $a_{s*}$ for space-, $a_{p*}$ population-related assumptions and $a_{*}$ for mixed assumptions. 

\subsection{Assumptions $\mathcal{A}^{t_0}(\mathcal{M}^{t_0})$}
\label{sec:generalform:initialAssumptions}

\underline{$\mathcal{A}^{t_0} = \{a^0_1, a^0_2, a^0_3, \dots, a^0_r\}$} is a finite set of assumptions concerned only with initial assumptions at the initial model state at time $t_0$. All equations from the previous sections apply only to the initial model state $M^{t_0}$. For instance, Equation \ref{eq:assumptionSatisfy} maps to 
\begin{equation}
\label{eq:t0assumptionSatisfy}
a^0_i \vdash \mathcal{M}^t_0  ~~~,~~~ \forall i \in \{ 1,2,\dots,r \} 
\end{equation}
Examples of initial assumptions are listed as follows:

\begin{center}
\graybox{
All adult persons have no parents
\begin{equation}
\label{eq:ass:adultHasNoParents}
    p \in P_{age \geq 18}^{t_0} \implies \nexists q \in P^{t_0} \text{ s.t.\ } q \in parents(p)
\end{equation}
}
%
%
\graybox{
All children have alive parents, i.e. 
\begin{equation}
\label{eq:ass:noOrphanChild} 
p \in P_{age<18}^{t_0} \implies parents(p) \subset P_{isAlive}^{t_0}
\end{equation}
}
\end{center}
with initial assumptions labeled as $a^0_{*}$.

\section{Single-clocked fixed-step simulation process}
\label{sec:abmsimulation}

Given Equation \eqref{eq:abmDefinition} expressing the model associated with the features, events, simulation parameters and assumption together with the ABM given in Equations \eqref{eq:abmModel} and \eqref{eq:abmModelInitialConditions}, an agent-based simulation process follows the pattern:
\begin{equation}
\label{eq:evolution}
\sum_{t=t_0}^{t_{final}-\Delta t} \prod_{i=1}^{n}  e_i ( \mathcal{M}^{t} )  = \mathcal{M}^{t_{final}}
\end{equation}
Illustratively, the evolution of the initial population and the corresponding initial featured sub-populations is defined as a sequential application of the events transitions: 
\begin{align*}
& \mathcal{M}^{t_0} & \text{ evaluating } & (P(t_0), P_f(t_0),S(t_0)) ~ \forall f \in \mathcal{F} 
& \xRightarrow{\mathcal{E}} \\  
& \mathcal{M}^{t_0 + \Delta t} &  \text{ evaluating } &  (P(t_0 + \Delta t), P_f(t_0 + \Delta t), S(t_0+ \Delta t)) ~ 
& \xRightarrow{\mathcal{E}} \\  
& \mathcal{M}^{t_0 + 2 \Delta t} &  \text{ evaluating } &  (P(t_0 + 2\Delta t), P_f(t_0 + 2\Delta t), S(t_0 + 2\Delta t) ) ~
& \xRightarrow{\mathcal{E}} \\ 
& \dots & \dots \\ 
& \mathcal{M}^{t_{final}} & \text{ evaluating } & (P(t_{final}), P_f(t_{final}), S(t_{final}))& 
\end{align*}
The initial model state $\mathcal{M}^{t_0}$ and the model states $\mathcal{M}^{[t_0,t_{final}]_{\Delta t}}$ should not violate the given assumptions $\mathcal{A}^{t_0}$ and $\mathcal{A}$ according to Equations \eqref{eq:t0assumptionSatisfy} and \eqref{eq:assumptionSatisfy}.

\else 
\section{Overview}
\label{sec:examplemodel}

In this and the following sections, a model example is introduced according to the specification terminology proposed in \cite{Elsheikh2023a}. Although, the document attempts to demonstrate an example of a model specification in a stand-alone manner, it is definitely helpful for a reader not familiar with the employed terminology to consult the cited article for in-depth clarification. \\ 

This section provides a brief overview while detailed specification are provided in the following sections. The brief overview is attempted to be sufficient for a general understanding of the model. The later detailed sections are appropriate for reproducing or exploiting the implementation.

\subsection{Description and aims}
\label{sec:example:informalOverview}

The model is concerned with a simplified demographic-only version of the lone parent model introduced in \cite{Gostoli2020}.
The presented model evolves an initial artificial population of the UK through a combination of events: births, deaths, marriages, divorces and ageing. The main purpose of the model is to act as 
\begin{itemize}
    \item 
    an experimental model for examining the capabilities of agent-based modeling libraries in the context of demographic modeling s.a.\ the Agents.jl package \cite{Datseris2022} 
    \item
    as a base for implementing further demographic-related case studies s.a.\ epidemiology or immigration among others   
\end{itemize}

\subsection{Formal overview}
\label{sec:example:formalOverview}

The specification of the model and its simulation process is basically established by describing the elements of tuple listed in Equation (22) in \cite{Elsheikh2023a}, namely:
\begin{equation}
    \tag{22}     
    < 
    \alpha_{sim}, 
    \mathcal{F}, 
    \mathcal{M}, \mathcal{M}^{t_0},  
    \mathcal{E}, 
    \mathcal{A}, \mathcal{A}^{t_0} 
    >
\end{equation}
with 
\begin{itemize}
    \item
    \underline{$\alpha_{sim} = (\Delta t, t_0, t_{final}, seed)^T$}: simulation parameters, cf.\ Section \ref{sec:parameters:simpar}
    \item 
    \underline{$\mathcal{F}$}: the set of employed population features, cf.\ Section \ref{sec:example:populationFeatures}
    \item 
    \underline{$\mathcal{M}$}:
    the given ABM model mathematical form, briefly overviewed in Section \ref{sec:example:model} and formulated in details in Section \ref{sec:model}
\item 
    \underline{$\mathcal{E}$}: the set of events, specified as
    \begin{equation}
    \label{eq:example:events}
    \mathcal{E} = \{ageing, births, deaths, divorces, marriages \}
    \end{equation}
    where detailed specification of each event is provided in Section \ref{sec:events}
    \item 
    \underline{$\mathcal{A}$}: the set of model assumptions that should not be violated are excessively summarized in Section \ref{sec:example:assumptions}
    \item \underline{$\mathcal{M}^{t_0}$ and $\mathcal{A}^{t_0}$} 
    the initial model states at the initial simulation time $t_0$ together with the initial model assumptions are detailed in Section \ref{sec:initialization}
\end{itemize}

\subsection{Simulation parameters $\alpha_{sim}$}
\label{sec:parameters:simpar}

In version 1.1 of the package MiniDemographicABM.jl \cite{MiniDemographicABMjlMisc}, Table \ref{tab:simparameters} lists selected default values of the simulation parameters out of other possible values: 
\begin{table}[h]
    \centering
\begin{tabular}{|l|c|c|} 
\hline 
$\alpha_{x}$ & default value & possible values \\ 
\hline 
$t_0$ & 2020 &  the first of January of 1800-2020 \\ 
$\Delta_t$ & Daily & \{Hourly, Daily, Monthly\} \\ 
$t_{final}$ & 2030 & the end of December of 2020-2100 \\ 
$seed$ & random & arbitrary \\ 
\hline
\end{tabular}
    \caption{Values of simulation parameters}
    \label{tab:simparameters}
\end{table}

\comment{It is beneficial in future to further propose several case studies with specific simulation parameter values for each case.}

\subsection{Population features $\mathcal{F}$}
\label{sec:example:populationFeatures}

The employed elementary population features are summarized Table \ref{Table:PopFeatures}.

\begin{table}[h]
\label{Table:PopFeatures}
    \centering
    \begin{tabular}{||c|c||}
    \hline
    \hline 
     Features &  Predicates \\
    \hline 
    \hline 
    age & $adult?$ ($ ~\equiv~ \neg child? ~\equiv~ age \geq 18?$  ), $age = a?$ \\ 
    \hline 
    status & $alive?$, $female?$, $male?$,  $orphan?$  \\ & $marriageEligible?$, $married?$, $single?$  \\ 
     & $gaveBirth?$, $oldestSibling?$, $reproducible?$ \\ 
     \hline 
    space & $empty?$, $house$, $livesAlone?$ \\ & $occupants$, $w?$ (i.e.\ in a town $w$) \\ 
    \hline 
    kinship &  $father$, $firstDegRelatives$, $mother$, $parents$ \\ 
            & $partner$,  $youngestChild$, $wife$ \\ 
    \hline 
    \hline 
    \end{tabular}
    \caption{Set of employed elementary population features}
    \label{tab:populationFeatures}
\end{table}
\todo{Complete}

\subsection{The model $\mathcal{M}$ and its initial state $\mathcal{M}^{t_0}$}
\label{sec:example:model}

Equations (23) and (24) in \cite{Elsheikh2023a}
\begin{align}
\label{eq:abmModel}
\tag{23}
& \mathcal{M}  \equiv \mathcal{M}^t \equiv \mathcal{M}(P(t),S(t),\alpha,D(t),t)   \\
\label{eq:abmModelInitialConditions}
\tag{24}
& \text{ associated with } \mathcal{M}(P(t_0),S(t_0),\alpha,D(t_0),t_0) =  M^{t_0} 
\end{align}
provide a brief description of the entities of the model where:
\begin{itemize}
    \item 
    $P \equiv P(t)$ corresponds to a population of individuals at time $t$
    \item 
    The space 
    \begin{equation}
    \label{eq:space}
    S ~\equiv~ S(t) ~=~ <H(t)~,~W>     
    \end{equation}
    corresponds to a dynamic set of houses $H(t)$ distributed within a static set of towns $W$ of the UK, cf.\ Section \ref{sec:model:space} for detailed description
    \item 
    model parameters $\alpha$ are provided in Section \ref{sec:model:parameters}
    \item 
    model input data $D$ is demonstrated in Section \ref{sec:model:data}
    \item 
     the initial model state $M^{t_0}$ is described in details in Section \ref{sec:initialization}
\end{itemize}

\subsection{Model assumptions $\mathcal{A}$ and initial assumptions $\mathcal{A}^{t_0}$} 
\label{sec:example:assumptions}

There are couple of distinguished subsets of assumptions: 
\begin{itemize}
    \item population-based assumptions (to be labeled with \boldmath$a_p$) 
    \item space-based assumptions (to be labeled with $a_s$)
    \item mixed (to be labeled with $a$) or 
    \item initial assumptions $a^{0}$
\end{itemize}
A summary of all model assumptions is given in Table \ref{tab:assumptions}. Detailed possibly formal description of the assumptions are distributed throughout the documentation whenever relevant to the context.

\begin{table}[t]
    \centering
    \begin{tabular}{|c|l|c|}
    \hline
    \textbf{label} & \textbf{summary}  & \textbf{context} \\
    \hline 
    \multicolumn{3}{|c|}{\textbf{initial population related assumptions} \boldmath$a_{*}^0$} \\
    \hline 
    \initassnum{\ref{ass:init:adultsNoParents}} & adults have no parents & Sec.\ \ref{sec:initialization:parents} \\
    \initassnum{\ref{ass:init:childrenAliveParents}} & parents are alive & " \\ 
    \initassnum{\ref{ass:init:ArbitraryAgeDiffSiblings}} & arbitrary age difference among siblings & " \\
    \initassnum{\ref{ass:init:familyTogether}} & family lives together & Sec.\ \ref{sec:initialization:housing} \\
    \hline 
    \multicolumn{3}{|c|}{\textbf{space-related assumptions} \boldmath$a_{s_*}$} \\ 
    \hline 
    \sassnum{\ref{ass:space:staticTowns}} & static set of towns & Sec.\ \ref{sec:model:space} \\ 
    \sassnum{\ref{ass:space:inhabitable}} & a just created house remains forever & " \\
    \sassnum{\ref{ass:space:dynamicSpace}} & dynamic space & " \\
    \sassnum{\ref{ass:space:dynamicHouses}} & dynamic set of houses & " \\
    \sassnum{\ref{ass:space:houseXYTown}} & town-xy-coordinate house location & " \\ 
    \sassnum{\ref{ass:space:housingUniformLocations}} & town houses uniformly distributed & " \\
    \sassnum{\ref{ass:space:selectOrCreateEmptyHouse}} & empty house selection (for new owners) & " \\ 
    \sassnum{\ref{ass:space:houseInArbitraryTown}} & population density oriented town selection & " \\ 
    \hline 
        \multicolumn{3}{|c|}{\textbf{population-related assumptions} \boldmath$a_{p_*}$} \\ 
    \hline 
   \passnum{\ref{ass:pop:gender}} & equal gender ratio & Sec.\ \ref{sec:initialization:gender} \& \ref{sec:events:births} \\  
   \passnum{\ref{ass:pop:marrigeaAge}} & only adults are married & Sec.\ \ref{sec:initialization:partnership} \& \ref{sec:events:marriages} \\  
    \passnum{\ref{ass:pop:marriedGivesBirth}} & only married non-old women give births &  Sec.\ \ref{sec:initialization:parents} \& \ref{sec:events:births} \\ 
   \passnum{\ref{ass:pop:adoption}} & no adoption for orphans & Sec.\ \ref{sec:events:deaths} \\
   \hline 
    \multicolumn{3}{|c|}{\textbf{mixed population-space related  assumptions} \boldmath$a_{*}$} \\ 
    \hline 
    \assnum{\ref{ass:mixed:homeless}}  & no homeless & Sec.\ \ref{sec:initialization:housing} \\
    \assnum{\ref{ass:mixed:numOfOccupants}} & arbitrary number of occupants & " \\
    \assnum{\ref{ass:mixed:housingKinship}} & flatmates are relatives & Sec.\ \ref{sec:events:order} \\ 
    \assnum{\ref{ass:mixed:adultToOwnHouse}} & new adults move out & Sec.\ \ref{sec:events:ageing} \\ 
    \assnum{\ref{ass:mixed:deadHaveNoHouse}}& deads leave their house & Sec.\ \ref{sec:events:deaths} \\ 
    \assnum{\ref{ass:mixed:divorceMaleToOwnHouse}} & divorced male moves out & Sec.\ \ref{sec:events:divorces} \\
    \assnum{\ref{ass:mixed:marriedHousing}} & housing assignment of new couple & Sec.\ \ref{sec:events:marriages} \\ 
    \hline 
    \end{tabular}
    \caption{Summary of model assumptions}
    \label{tab:assumptions}
\end{table}
\section{The model $\mathcal{M}$} 
\label{sec:model}

\subsection{The space $S$}
\label{sec:model:space}

\subsubsection{Space-related model assumptions $a_{s_*}$} 

Before diving into the specification of the space $S$, it makes sense to list some related set of space-oriented assumptions. The space $ S $, cf.\ Equation \eqref{eq:space} is composed of a tuple $<H(t)~,~W>$  corresponding to the set of all houses $H(t)$ and towns $W$ implying that: 
\grayspass{
\label{ass:space:staticTowns} The set of towns is constant during a simulation, i.e.\ no town vanishes nor new ones get constructed:
\begin{equation}
\label{eq:ass:staticTowns}
    \text{ if } t_0 \leq t_1 \neq t_2 \leq t_{final} \implies W(t_1) \equiv W(t_2) \equiv W   
\end{equation}
}
\grayspass{
\label{ass:space:inhabitable}
Once a new house is built (e.g.\ to host an adult moving out of his parent's house), it never gets demolished and remains always inhabitable 
\begin{equation}
\label{eq:ass:HouseAlywasRemain}
    \text{ if } h \in H(t) ~\implies h \in post(H(t)) 
    \text{ where } t_0 \leq t < t_{final}
\end{equation}
}
Based on the space definition and the previous Equation, the following assumption is considered:
\grayspass{
\label{ass:space:dynamicSpace}
The space is not necessarily static and particularly the set of houses can vary along the simulation time span, i.e.
\begin{equation}
\label{eq:dynamicHouses}
\text{ if } t_0 \leq t < t_{final} \implies  H(t) \subseteq post(H(t)) 
\end{equation}
}
Consequently,
\grayspass{
\label{ass:space:dynamicHouses}
Each town $w \in W$ contains a dynamic set of of houses
\begin{equation}
\label{eq:ass:dynamicSetOfHousesInATown}
w(H^t)  \equiv  H_w^t \equiv H_w(t)  
\end{equation}   
}

\subsubsection{The space $S$ -- description}

In the sake of simplifying the implementation of the space,  
the static set of towns of UK, cf.\ \sassnum{\ref{ass:space:staticTowns}}, is projected as a rectangular $12 \times 8$ grid with each point in the grid corresponding to a town \cite{Gostoli2020}.
%
Formally, assuming that 
$$location(w_{(x,y)}) = (x,y) $$ 
then
\begin{itemize}
\item the town $w_{(1,1)}$ corresponds to the north-est west-est town of UK whereas 
\item the town $w_{(12,8)}$ corresponds to the south-est east-est town of UK 
\item  the distances between towns are commonly defined, e.g.\ 
\begin{equation}
\label{eq:manhattan}
\text{manhattan-distance}(w_{(x_1,y_1)} , w_{(x_2,y_2)}) = \mid x_1 - x_2 \mid + \mid y_1 - y_2 \mid
\end{equation}
\end{itemize}

The (initial) population and houses distribution within UK towns are approximated by an ad-hoc pre-given UK population density map.
The map is projected as a rectangular matrix  
\begin{equation}
\label{eq:M}
M  \in R^{12 \times 8} \approx 
\begin{bmatrix} 
0.0 & 0.1 & 0.2 & 0.1 & 0.0 & 0.0 & 0.0 &  0.0 \\
0.1 & 0.1 & 0.2 & 0.2 & 0.3 & 0.0 & 0.0 & 0.0 \\
0.0 & 0.2 & 0.2 & 0.3 & 0.0 & 0.0 & 0.0 & 0.0 \\
0.0 & 0.2 & 1.0 & 0.5 & 0.0 & 0.0 & 0.0 & 0.0 \\ 
0.4 & 0.0 & 0.2 & 0.2 & 0.4 & 0.0 & 0.0 & 0.0 \\
0.6 & 0.0 & 0.0 & 0.3 & 0.8 & 0.2 & 0.0 & 0.0 \\ 
0.0 & 0.0 & 0.0 & 0.6 & 0.8 & 0.4 & 0.0 & 0.0 \\ 
0.0 & 0.0 & 0.2 & 1.0 & 0.8 & 0.6 & 0.1 & 0.0 \\ 
0.0 & 0.0 & 0.1 & 0.2 & 1.0 & 0.6 & 0.3 & 0.4 \\ 
0.0 & 0.0 & 0.5 & 0.7 & 0.5 & 1.0 & 1.0 & 0.0 \\ 
0.0 & 0.0 & 0.2 & 0.4 & 0.6 & 1.0 & 1.0 & 0.0 \\ 
0.0 & 0.2 & 0.3 & 0.0 & 0.0 & 0.0 & 0.0 & 0.0
\end{bmatrix}
\end{equation} 

It can be observed for instance that
\begin{itemize}
\item 
cells with density $0$ (i.e.\ realistically, with very low-population density) don't correspond to inhabited towns  
\item 
the towns in UK are merged into 48 towns
\item e.g. the center of the capital London spans the cells $(10,6), (10,7), (11,6)$ and $ (11,7)$ 
\end{itemize}

\subsubsection{Further space-related assumptions}

Further assumptions are needed for specification of houses, their creation and their distributions within the UK: 
\grayspass{
\label{ass:space:houseXYTown}
The static location of a house $h \in H^{[t,t_{final}]}_w$ is given in xy-coordinate of the town
\begin{equation}
location(h) = (x,y)_w \text{ where } 1 \leq x,y \leq 25    
\end{equation}
}
\grayspass{
\label{ass:space:housingUniformLocations}
The locations of houses $H_w$ within a town $w \in W$ are uniformly distributed along the x- and y- axes  
\begin{equation}
\label{eq:uniformDistHouses}
Dist(location(H_w)) \propto   (U^{[1,25]},U^{[1,25]})
\end{equation}
}
The previous equation reads the distribution is proportional to. Furthermore,
\grayspass{
\label{ass:space:selectOrCreateEmptyHouse}
If an empty house $h$ is demanded in a particular town $w \in W$, an empty house is randomly selected from the set of existing empty houses $H_{w(empty)}$ in that town $w$
\begin{equation}
\label{eq:}
h_w = random(H_{w(empty)}) 
\end{equation}
If no empty house exists, a new empty house is established according to assumptions \sassnum{\ref{ass:space:housingUniformLocations}}
and \sassnum{\ref{ass:space:houseXYTown}}
}
\grayspass{
\label{ass:space:houseInArbitraryTown}
If an empty house $h$ is demanded in an arbitrary town, a town is selected via a random weighted selection, say: 
\begin{align}
\label{eq:ass:hometowLocation}
 town(h) =  random(W,M) 
\end{align}
an empty house is selected or established according to the previous assumption 
}

\subsection{Model parameters $\alpha_*$}
\label{sec:model:parameters}

The following is a table of parameters employed for events specification, cf.\ Section \ref{sec:events}. The values are set in an ad-hoc manner as they are not calibrated to actual data. The choice of data rather depends on the simulation parameters, e.g.\ the start and final simulation times, as well as the underlying case study.
\begin{center}
\begin{tabular}{|l|c|l|} 
\hline 
$\alpha_{x}$ & Value & Usage  \\
\hline 
$basicDivorceRate$ & 0.06 & Equation \ref{eq:prob:divorce} \\ 
$basicDeathRate$ & 0.0001 & Equation \ref{eq:probDeath} \\ 
$basicMaleMarriageRate$ & 0.7 & Equation \ref{eq:prob:marriage} \\ 
$femaleAgeDeathRate$ & 0.00019 & Equation \ref{eq:probDeath} \\ 
$femaleAgeScaling$ & 15.5 & Equation \ref{eq:probDeath} \\
$initialPop$ & 10000 & Section \ref{sec:initialization:initialPopulation} \\ 
$maleAgeDeathRate$ & 0.00021 &  Equation \ref{eq:probDeath} \\ 
$maleAgeScaling$ & 14.0 & Equation \ref{eq:probDeath} \\ 
$maxNumMarrCand$ & 100 & Sections \ref{sec:initialization:partnership} \& \ref{sec:events:marriages} \\
${startMarriedRatio}$ & 0.8 & Equation \ref{eq:initialMarriedProb} \\
\hline
\end{tabular}
\end{center}
The value of the initial population size is just an experimental value and can be selected, for instance, from the set $\left\{ 10^4, 10^5, 10^6 , 10^7, 10^8 \right\}$ to examine the runtime performance of specific implementation and/or whether it is possible to enable a realistic demographic simulation with an actual population size.


\subsection{Input data $D(t)$}
\label{sec:model:data}

In the archived Julia package MiniDemographicABM.jl \cite{MiniDemographicABMjlMisc}, fertility data is given as: 
\begin{align*}
& D_{fertility} \in R^{35 \times 360} ~= \nonumber \\ 
& ~~~~~ \left[ d_{ij} : \text{ fertility rate of women of age } i-16 \text{ in year } j - 1950   \right]    
\end{align*}
This matrix, taken from the Python implementation of the Lone Parent Model \cite{Gostoli2020}, reveals (the forecast of) the fertility rate for woman of ages 17 till 51 between the years 1951 and 2050, cf.\ Figure \ref{fig:fertilityData}. 
\begin{figure}[h]
\centering
\includegraphics[width=4in]{fig/fertility.png}
\caption{Fertility rates of women of different age classes between the years 1966 and hypothetically till 2310}
\label{fig:fertilityData}
\end{figure}

Furthermore, the following (ad-hoc) data values, subject to tuning,: 
\begin{align*}
   & D_{divorceModifierByDecade} \in R^{16} ~= \nonumber \\ 
   &~~~~~ (0,1.0,0.9,0.5,0.4,0.2,0.1,0.03,0.01,0.001,0.001,0.001,0,0,0,0)^T  
\end{align*}
\begin{align*}
   & D_{maleMarriageModifierByDecade} \in R^{16}  ~= \nonumber \\ 
   &~~~~~ (0,0.16,0.5,1.0,0.8,0.7,0.66,0.5,0.4,0.2,0.1,0.05,0.01,0,0,0)^T   
\end{align*}
are employed in event specification, cf.\ Equations \eqref{eq:prob:divorce} and \eqref{eq:prob:marriage} for their contextual interpretation.


\section{Model initialization $\mathcal{M}^{t_0}$} 
\label{sec:initialization}

\setvalue{\popassnum}{6} \todo{to remove}


\subsection{Initial population size and distribution $|P^{t_0}_{w}|$}  
\label{sec:initialization:initialPopulation} 

The initial population size is given by the parameter $\alpha_{initialPop}$, cf.\ Section \ref{sec:model:parameters}. 
The matrix $M$ and assumption \sassnum{\ref{ass:space:houseInArbitraryTown}} provide together a stochastic ad-hoc estimate of the initial population distribution within the UK. 
That is, the initial population size of a town $w \in W$ is estimated as
\begin{equation}
\label{eq:Pwt0}
|P_w(t_0)| \approx \left\lceil \alpha_{initialPop} \times M_{y,x} / 48 \right\rceil  \text{~~ where ~~location}(w) = (x,y) 
\end{equation}
where 48 is the number of nonzero entries in $M$.

\subsection{Gender $P^{t_0} = M^{t_0} \cup F^{t_0}$}  
\label{sec:initialization:gender} 

The gender ratio distribution is specified via the following (clearly non-realistic) population-related assumption 
%
\graypopass{
\label{ass:pop:gender}
An individual can be equally a male or a female, i.e.
\begin{equation}
\label{eq:genderProbability}
Pr(p \in P^{t_0}_{male}) \approx 0.5
\end{equation}
}
This assumption is employed in the specification of the initial population as well as the specification of the birth event, cf.\ Section \ref{sec:events:births}, and this it is not classified as an initial assumption. 


\subsection{Age distribution $P^{t_0} = \bigcup_{r} P_{age=r}^{t_0}$}

The proposed non-negative age distribution of population individuals in years follows a normal distribution: 
\begin{equation}
\label{eq:ageDistribution}
Dist \left( \frac{age(P^{t_0})}{N_{\Delta t}} \in \mathbb{Q}^{\alpha_{initialPop}}_+ \right) ~\propto~ \left| \lfloor \mathcal{N}(0,  \frac{100}{4} \cdot N_{\Delta t}) \rfloor \right|
\end{equation}
where $\mathbb{Q}_+$ stands for the set of positive rational numbers and  $\mathcal{N}$ stands for a normal distribution with mean value 0 and standard deviation depending on 
\begin{equation}
\label{eq:NDeltaT}
    N_{\Delta t} ~=~ \left\{ 
\begin{array}{ll}
~ \dots \\
~ 12 &  \text{ if }~ \Delta t = month \\ 
365 & \text{ if }~ \Delta t = day \\ 
365 \cdot 24 & \text{ if }~ \Delta t = hour \\ 
~ \dots 
\end{array}
\right. 
\end{equation}
A possible outcome of the distribution of ages in an initial population of size 1,000,000 is shown in Figure \ref{fig:initialPopulationAges}.
\begin{figure}[h]
\centering
\includegraphics[width=3in]{fig/initpop.png}
\caption{The distribution of ages in an initial population of size 1,000,000}
\label{fig:initialPopulationAges}
\end{figure}
Obviously, in reality, the shape of the age's distribution depends on the initial simulation time $t_0$.

\subsection{Partnership $P_{married}^{t_0} = M_{married}^{t_0} \cup partner(M_{married}^{t_0})$}\todo{Check this}
\label{sec:initialization:partnership}

Initially the following population assumption concerned with marriage age is considered
\graypopass{
\label{ass:pop:marrigeaAge} 
A married person is an adult person  
\begin{align}
p \in P_{married} \implies age(p) \geq 18 
\end{align}
}

%
The ratio of married adults (males or females) is statistically approximated according to 
\begin{equation}
\label{eq:initialMarriedProb}
Pr(p \in P_{isSingle ~\cup~ \neg adult }^{t_0}) \approx 1 - 2 \cdot \alpha_{startMarriedRatio} 
\end{equation}
That is, $\alpha_{startMarriedRate}$ is the ratio of married males (females) among adults. 
Partnership initialization is established according to Algorithm \ref{alg:partnerInitialization}. Initially lines 1-3 initialize \begin{enumerate}
    \item the set of males randomly selected for marriage (line \ref{line:alg:partner:line1})
    \item the set of females eligible to marriage (line 2)
    \item  and the number of female candidates for marriage each male has to select from\footnote{This is just an abstract algorithm that does not necessarily reflect the reality} (line 3). 
\end{enumerate} 
For every male selected for marriage, a corresponding female is selected (lines 4-11): 
\begin{enumerate}
    \item a set of candidate females is initialized (line 5) 
    \item for every candidate female (line 6), a weight is calculated according to a weight function based on age difference in Equation \eqref{eq:marriageWeight} (line 7) 
    \item 
    a female partner is selected according to a random weighted function (line 9) 
    \item 
    the set of females illegible for marriage is updated (line 10)
\end{enumerate}
Using the calculated weights, a partner 
\begin{algorithm}
\label{alg:partnerInitialization}
\caption{Partnership initialization}
\begin{algorithmic}[1]
\State \label{line:alg:partner:line1}
Select $M_{married}$ randomly according to Equation \eqref{eq:initialMarriedProb} 
\State
Set $F_{marriageEligible}^{t_0} = F_{marEli}^{t_0} = F_{adult ~\cap~ single}^{t_0}$
\State 
Set $n_{candidates} = max\left( \alpha_{maxNumMarrCand} ~,~ \frac{|F_{marEli}|}{10}\right)$ 
\For{$m \in M_{married}$} 
\State Set $ F_{candidates} = random(F_{marEli}, n_{candidates})$ 
    \For{$f \in F_{candidates}$} 
        \State \label{line:alg:partner:weightFunc} Set $w_{m,f} = weight(m,f) = ageFactor(m,f)$ where
    \begin{align}
    \label{eq:alg:ageFactor}
        & ageFactor(m,f)  =~  \nonumber \\ & ~~~ \left\{ 
\begin{array}{ll}
~ 1 / (age(m) - age(f) - 5 + 1) &  \text{ if }~ age(m) - age(f) \geq 5 \\ 
~ -1 / (age(m) - age(f) + 2 - 1) &  \text{ if }~ age(m) - age(f) \leq -2 \\ 
~ 1 \text{ ~~~~otherwise } \\ 
\end{array}
\right.   
\end{align}
    \EndFor
\State Set 
\begin{align}
& f_{partner(m)}  =  weightedSample(F_{candidates}, W_m) 
\nonumber \\ 
& ~~~~~ \text{ where } ~ W_m = \left\{ w_i : w_i = weight(m,f_i) ~,~ f_i \in F_{marEli} \right\}     
\end{align}
\State 
Set $F_{marEli} = F_{marEli} - \{ f_{partner(m)}\}$
\EndFor 
\end{algorithmic}
\end{algorithm}

\subsection{Children and parents} 
\label{sec:initialization:parents}

The following assumptions are assumed only in the context of the initial population: 
%
%
\grayinitass{
\label{ass:init:adultsNoParents}
All adult persons have no parents
\begin{equation}
\label{eq:ass:adultHasNoParents}
    p \in P_{age \geq 18}^{t_0} \implies \nexists q \in P^{t_0} \text{ s.t.\ } q \in parents(p)
\end{equation}
} 
%
\grayinitass{
\label{ass:init:childrenAliveParents}
All children have alive parents, i.e. 
\begin{equation}
\label{eq:ass:noOrphanChild} 
p \in P_{age<18}^{t_0} \implies parents(p) \subset P_{alive}^{t_0}
\end{equation}
}
Based on the previous two assumptions, one can may also deduce that there is no individual in the initial population who has a grandpa or grandma. 
\grayinitass{
\label{ass:init:ArbitraryAgeDiffSiblings}
There is no age difference restriction among siblings, i.e.\ age difference can be less than 9 months
}
Moreover, the following population assumption proposes conditions for women who can give birth:
\graypopass{
\label{ass:pop:marriedGivesBirth}
Only a married female\footnote{This was assumed in the lone parent model and obviously the marriage / partnership concept needs to be re-defined in the context of realistic studies} under age of 45 gives birth
\begin{align}
\label{eq:ass:pop:onlyMarriedGivesBirth}
f \in & F_{just(gaveBirth)}^t \implies \nonumber \\ & 
f \in F_{married}^t \text{ and } age(f) < 45
\end{align}
}
Assumptions \passnum{\ref{ass:pop:marriedGivesBirth}} and \passnum{\ref{ass:pop:marrigeaAge}} imply that only an adult person can become a parent. 
Children are assigned to married couples as parents in the following way. 
For any child $c  \in P_{age<18}^{t_0}$, the set of potential fathers is established as follows: 
\begin{align}
\label{eq:fathershipCanddiates}
M_{candidates} = ~ & \{~ m \in M_{married} \text{ s.t.\ } \nonumber  \\ &   min\left(~age(m),age(wife(m))~\right) \geq age(c) + 18 + \frac{9}{12} \text{ and } \nonumber \\ & 
age(wife(m)) < 45 + age(c) ~\}  
\end{align}
out of which a random father is selected for the child: 
\begin{align*}
    father(c) =  & random(M_{candidates}) \text{ and } \\
    & mother(c) = wife(father(c)) 
\end{align*}

\subsection{Housing assignments}
\label{sec:initialization:housing}

Before specifying the housing assignments to initial population, related model assumptions are listed:  
\grayass{
\label{ass:mixed:homeless} There are no homeless individuals:
\begin{equation}
\label{eq:mixed:pop:noHomeless}
    \text{ if } p \in P^t_{alive} ~\implies~ house(p^t) \in H(t)
\end{equation}
}
The last line in the previous equations indicates that a dead person does not need to be associated to any house. 
Furthermore, there is no classification of houses according to their capacities:
\grayass{
\label{ass:mixed:numOfOccupants}
A house can be occupied by an arbitrary number of individuals
}
Moreover, the following assumption is considered for housing association:
\grayinitass{
\label{ass:init:familyTogether}
A family, i.e.\ a married male and female, and their children live together, that is 
\begin{align}
\label{eq:ass:familyTogether}
    |occupants(h^{t_0})| > 1 & \text{ with } p,q \in occupants(h^{t_0}) \text{ and } p \neq q ~\implies~ \nonumber \\ &  p \in firstDegRelatives(q)  
\end{align}
} 
The previous assumption implies that a single person in the initial population shall be assigned a house alone.  
The assignment of newly established houses to initial population considers the assumptions \sassnum{\ref{ass:space:selectOrCreateEmptyHouse}} and \sassnum{\ref{ass:space:houseInArbitraryTown}}. That is, 
the location of new houses in $H(t_0)$ is specified according to to Equation \eqref{eq:ass:hometowLocation}, each is assigned to a single person or a family as previously stated. 
%

\section{Events}
\label{sec:events}

This section provides compact algorithmic specification of events serving as a comprehensive demonstration of the proposed terminology presented in \cite{Elsheikh2023a}.   

\subsection{Execution order of events}
\label{sec:events:order}

Despite Equation \eqref{eq:example:events} specifying the set of events, the considered events are just alphabetically listed without enforcing a certain appliance order, except for the ageing event which should proceed any other events.
That is,
$$e_1 = ageing$$ 
This is reasonable since if an event s.a.\ death or birth proceeds ageing, then this implies that the population size and features may not remain consistent with input data $D(t)$, model parameters $\alpha$ and initial model states $\mathcal{M}^{t_0}$. \\ 

The execution order of the rest of the events as well as the order of the agents subject to such events, whether sequential or random, remains an implementation detail.  
Nevertheless, since many of the events are following a random stochastic process, probably, the higher the resolution of the simulation becomes (e.g.\ weekly step-size instead of monthly, or daily instead of weekly), the less influential the execution order of the events becomes. \\ 

The combination of event transitions on the population does not preserve the initial assumption \initassnum{\ref{ass:init:familyTogether}} regarding the occupants of houses. Therefore, assumption regarding the occupants of houses is further relaxed to: 
\grayass{
\label{ass:mixed:housingKinship} Any two individuals living in a single house are either a 1-st degree relatives, step-parent, step-child, step-siblings or partners. An exception to the previous assumption occurs when an orphan's oldest sibling is married in which case they also lives together as a family.  
}

\subsection{Ageing}
\label{sec:events:ageing}

The ageing process of a population can be described as follows:
\begin{equation}
\label{eq:event:ageing}
ageing \left(
P_{alive(age=a)}^t\right) 
= 
P_{alive(age=a+\Delta t)}^{t+\Delta t} ~~ ,~~ \forall a \in \{0, \Delta t, 2 \Delta t, ... \}
\end{equation}
That is, the age of any individual as long as he remains alive is incremented by $\Delta t$ for each simulation step. Furthermore, the following assumption concerned with individuals becoming adults is considered: 
\grayass{
\label{ass:mixed:adultToOwnHouse}
In case a teenager becomes an adult and he/she is not the oldest orphan, he/she gets re-allocated to an empty house in the same town.
}
Formally:
\begin{align}
\label{eq:AdultMovesToAnEmptyHouse}
ageing & \left(P_{alive(age = 18 - \Delta t) ~\cap~ \neg \left( orphan ~\cap~ oldestSibling \right) }  \right)  =  \nonumber \\ 
 & P_{alive(age = 18 ) ~\cap~ livesAlone } 
\end{align}
Moreover, 
\begin{align}
\label{eq:AdultMoveToTheSameTown}
 \text{If } p \in P^{t}_{alive(age=18)} \implies 
 town(p) = pre(town(p))   
\end{align}
The re-allocation to an empty house should be according to assumption \sassnum{\ref{ass:space:selectOrCreateEmptyHouse}}.

\subsection{Births}
\label{sec:events:births}

For simplification purpose, from now on, it is implicitly assumed (unless specified) that only the alive population is involved in event-based transition of population features. 
Given assumption \ref{ass:pop:marriedGivesBirth}, let the set of reproducible females be defined as:
\begin{align}
F_{reproducible} & ~=~ F_{married ~\cap~ age < 45} ~\bigcap~ \nonumber \\ &  F_{age(youngestChild) > 1 ~\cup~ \neg hasChildren} 
\end{align} 
That is, the set of all married females in a reproducible age and either do not have children or those with youngest child older than one.  
The birth event produces new children from reproducible females specified as follows
\begin{align}
birth & \left(F^t_{reproducible} \right) ~=~ \nonumber \\ 
&~  \left( F_{reproducible}^{t+\Delta t} ~ - ~ F_{just(reproducible)}^{t+\Delta t}  \right) ~ \bigcup  \nonumber \\ 
&~ F_{just(\neg reproducible) }^{t+\Delta t}  ~\bigcup \nonumber \\ 
 &~  P_{age=0}^{t+\Delta t}
\end{align} 
The previous equation states that the birth event transients the individuals within the set of reproducible females to: 
\begin{itemize}
\item those females who remained reproducible (first line in the rhs) 
\item those who just gave births (second line) and 
\item new neonates are produced (third line)  
\end{itemize}
Note that the following set is subtracted from those who remained reproducible 
\begin{equation}
F_{just(reproudicble)} ~=~ 
F_{just(married)} ~\cup~ 
F_{married(age(youngestChild)=1)}    
\end{equation}
and (given Assumption \passnum{\ref{ass:pop:marriedGivesBirth}}) those who became non-reducible include also who got divorced and widowed 
\begin{align}
& F_{just (\neg reproducible)} ~=~ \nonumber \\  
    & ~~~ F_{age(youngestChild)=0} ~ \cup ~ F_{just(\neg married)} ~\cup ~ F_{married(age=45)}    
\end{align}
%
%
Employing assumptions \assnum{\ref{ass:mixed:housingKinship}} and \assnum{\ref{ass:mixed:adultToOwnHouse}}, one can deduce that a neonate is assigned to his parents house, i.e. 
\begin{align}
p \in~ &  P_{age=0}   
\implies \nonumber \\ 
& house(p) = house(mother(p))    
\end{align}
The yearly-rate of births produced by the sub-population $F_{reproducible,(age=a)}^t$ i.e.\ reproducible females of age $a$ years old with actual simulation time $t$, depends on the yearly-basis fertility rate data:  
\begin{equation}
    R_{birth,yearly} (F_{reproducible,(age=a)}^t)  ~\propto~ D_{fertility}(a-16,currentYear(t))   
\end{equation}
This implies that the instantaneous probability that a reproducible female $f \in F_{reproducible}^t$ gives birth to a new individual $p \in P^{t + \Delta t}_{age=0}$ depends on $D_{fertility}(a-16,currentYear(t))$ and is given by Equation \eqref{eq:instantaneous}, cf.\ Appendix \ref{sec:terminology:probability} for a conceptual review regarding rates and instantaneous probabilities.

\subsection{Deaths} 
\label{sec:events:deaths}

The death event transforms a given population of alive individuals as follows: 
\begin{align}
\label{eq:event:deaths}
    death \left( P_{alive }^t \right)  ~=~  P_{alive - age=0}^{t+\Delta t}  ~\bigcup~  P_{just(\neg alive)}^{t+\Delta t} 
\end{align}
where
\begin{itemize}
    \item the first phrase in the right hand side stands for the alive population except neonates as they don't belong to $P^t_{alive}$
    \item the second phrase stands for those who just became dead
\end{itemize}
The following simplification assumptions are considered: 
\graypopass{
\label{ass:pop:adoption} 
No adoption or parent re-assignment to orphans is established after their parents die 
\begin{align}
    \label{eq:noAdoption}
    p \in P_{child} \text{ and }  parents(p) & \subset P_{\neg alive} \implies \nonumber \\ 
    & post(parents(p)) \subset P_{\neg alive}
\end{align}
} 
\grayass{
\label{ass:mixed:deadHaveNoHouse}
Those who just became dead they leave their houses, i.e.\ 
\begin{align}
\label{eq:deadsLeaveToGrave}
\text{ if } p \in P_{just(\neg alive)}^{t+\Delta t}~  &
\text{ and }~ pre(house(p)) = h 
\nonumber \\ 
\implies ~ & p \notin P^{t+\Delta t}_h  ~\text{ and }~ house(p) \notin H^{t + \Delta t}    
\end{align}
}
Note that $P_h$ stands for the population (or occupants) of a house $h$. 
The amount of population deaths depends on the yearly-rate given by:  
\begin{align}
\label{eq:probDeath}
R_{death,yearly}&(p \in P) = \alpha_{basicDeathRate} ~+~ \nonumber \\ & \left\{   
\begin{array}{cc}
   \left( e^{\frac{age(p)}{\alpha_{maleAgeScaling}}} \right)  \times \alpha_{maleAgeDeathRate}   \text{ if } male?(p) \\
       \left( e^{\frac{age(p)}{\alpha_{femaleAgeScaling}}} \right)  \times \alpha_{femaleAgeDeathRate}
    \text{ if } female?(p) \\
\end{array}
\right. 
\end{align}
from which instantaneous probability of the death of an individual is derived as illustrated in Appendix \ref{sec:probability}.

\subsection{Divorces} 
\label{sec:events:divorces}

The divorce event causes that a subset of married population becomes divorced: 
\begin{align}
& divorce (M_{married}^t) ~ = ~  \nonumber \\ 
 &~~~~~   M_{married - just(married)}^{t+\Delta t}  ~ \bigcup ~ M_{just(isDivorced)}^{t + \Delta t}  
\end{align}
The first phrase in the right hand side refers to the set of married individuals who remained married excluding those who just got married.
The second phrase refers to the population subset who just got divorced in the current iteration. 
Note that it is sufficient to only apply the divorce event to  either the male or female sub-populations. 
After divorce takes place, the housing's assignment is specified according to the following assumption: 
\grayass{
\label{ass:mixed:divorceMaleToOwnHouse}
Any male who just got divorced moves to an empty house within the same town (in conformance with Assumption \sassnum{\ref{ass:space:selectOrCreateEmptyHouse}}):
\begin{align}
 m \in & M_{just(isDivorced}^t) \implies \nonumber \\  
& pre(location(m)) \neq location(m) ~\text{ and }~ livesAlone(m) ~\text{ and }~ \nonumber \\  
& town(m) = pre(town(m))  
\end{align}
}
The re-allocation to an empty house is in conformance with assumption \sassnum{\ref{ass:space:selectOrCreateEmptyHouse}}.
The amount of yearly divorces in married male populations can be estimated upon the yearly divorce rate given by
\begin{align}
\label{eq:prob:divorce}
 R_{divorce,yearly}(m \in M_{single}^t) & ~=~ \nonumber \\ \alpha_{basicDivorceRate}  ~\cdot~ &  D_{divorceModifierByDecade}(\lceil age(m) / 10 \rceil)   
\end{align}
That is, the instantaneous probability of a divorce event to a married man $m \in M_{married}$ depends on $D_{divorceModifierByDecade}(\lceil age(m) / 10 \rceil)$, cf.\ Equation \eqref{eq:instantaneous}.

\subsection{Marriages}
\label{sec:events:marriages}

Similar to the divorce event, it is sufficient to apply the marriage event to a sub-population of single males. 
Assuming that 
\begin{equation}
\label{eq:MisMarEli}
M_{marEli} ~=~ M_{marriageEligible} ~=~ M_{single ~\cup~ age \geq 18}    
\end{equation}
the marriage event updates the state of few individuals within a sub-population to married males, formally: 
\begin{align}
\label{eq:event:marriage}
& marriage (M_{marEli}^t)  ~ = ~  \nonumber \\ 
 & ~~~~~   M_{marEli - just(isDivorced) - age=18}^{t+\Delta t}  
 ~ \bigcup ~ 
 M_{just(married)}^{t+\Delta t}  
\end{align} 
The amount of yearly marriages can be statistically estimated by
\begin{align}
\label{eq:prob:marriage}
& R_{marraige,yearly}(m \in M_{single}^t) ~=~  \nonumber \\ & ~ \alpha_{basicMaleMarriageRate}  ~\cdot~ D_{maleMarriageModifierByDecade}(\lceil age(m) / 10 \rceil)   
\end{align}
from which simulation-relevant instantaneous probability is calculated as given in Equation \eqref{eq:instantaneous}. 
For an arbitrary just married male $m \in M_{just(married)}^{t + \Delta t}$, 
his wife was selected according to a slight modification of Algorithm \ref{alg:partnerInitialization}. Namely line \ref{line:alg:partner:weightFunc} is modified to: 
\begin{align}
\label{eq:marriageWeight}
        & weight(m,f) ~=~ \nonumber \\  
        & ~~~ geoFactor(m,f) ~\cdot~ childrenFactor(m,f) ~\cdot~ ageFactor(m,f) 
    \end{align}
        \text{ and } 
    \begin{align}
        &        geoFactor(m,f)  =~ \nonumber \\ &~~~~~ 1 / e^{(4 \cdot \text{manhattan-distance}(town(m), town(f)))} \\ 
        & childrenFactor(m,f)  =~ \nonumber \\ 
        & ~~~~~ 1/e^{|children(m)|} \cdot 1/e^{|children(f)|} \cdot e^{|children(m)| \cdot |children(f)|}  
\end{align}
Where $ageFactor(m,f)$ is given in Equation \eqref{eq:alg:ageFactor}. Note that the just married male and his female partner don't then belong to the set of marriage eligible population $P_{marEli}^{t+\Delta 
t}$. \comment{It is thinkable to reverse the genders in the algorithm}
The following assumption specifies the housing's assignment of the new couple. 
\grayass{
\label{ass:mixed:marriedHousing}
When two individuals get married, the wife and the occupants of actual house (i.e.\ children and non-adult orphan siblings) moves to the husband's house unless there are fewer occupants in his house. In the later case, the husband and the occupants of his house move to the wife's house.  
}
Formally, suppose that $m \in P^{t+\Delta t}_{just(married)}$ if
\begin{align*}
 |P_{house(m^t)}| \geq & |P_{house(f^t)}|   \implies   house(p^{t+\Delta t}) = house(m), ~\forall p^t \in P_{house(f^t)}    
\end{align*}
Otherwise
\begin{align}
 house(p^{t+\Delta t}) = house(f),   ~\forall p^t \in P_{house(m^t)}      
\end{align}

\appendix
\section{Events rates and instantaneous probability}
\label{sec:terminology:probability}
\label{sec:probability}

Pre-given data, e.g. mortality and fertility rates, are usually given in the form of finite rates (i.e.\ cumulative rate) normalized by sub-population length. 
In other words, the rate 
$$R_{event, period(t,t+\Delta t)}(X^t) ~,~  X^t \subseteq P^t $$ 
corresponds to the number of occurrences that a certain $event$ within a sub-population $X$ (e.g.\ marriage) takes place in the time range between $(t,t+\Delta t)$, e.g.\ a daily, weekly, monthly or yearly rate, etc. normalized by the sub-population length. 
That is, say if a pre-given typically yearly rate is given as input data:
\begin{equation}
\label{eq:yearlyrateAsData}
R_{event,yearly}(X^t) ~=~ D_{event,yearly} \in R^{N \times M}
\end{equation} 
where $M$ corresponds to a given number of years and $N$ corresponds to the number of particular features of interest, for examples:
\begin{itemize}
    \item $M = 100$ for mortality or fertility yearly rate data between the years 1921 and 2020 
    \item $N = 28$ for fertility rate data for women of ages between 18 and 45 years old, i.e.\ $28 = 45-18+1$ 
\end{itemize}
The yearly probability that an event takes place for a particular individual $x^t \in X^t$ is:
\begin{equation}
\label{eq:yearlyProbability}
Pr_{event,yearly}(x^t \in X^t) ~=~ D_{event,yearly}(yearsold(x^t),currentYear(t)) 
\end{equation} 

Pre-given data in such typically yearly format desires adjustments in order to employ them within a single clocked agent-based-model simulation of a fixed step size typically smaller than a year. 
Namely, the occurrences of such events need to be estimated at equally-distant time points with the pre-given constant small simulation step size $\Delta t$. 
For example, if we have population of 1000 individuals with a (stochastic) monthly mortality rate of $0.05$, then after
\begin{itemize}
    \item one month (about) 50 individuals die with 950 left (in average)
    \item two months, about $902.5$  individuals are left
    \item $\dots$
    \item one year, 540 individuals are left resulting in a yearly finite rate of $0.46$
\end{itemize}
\comment{may be a figure}
\comment{better example based on daily rate , e.g. daily rate of 0.001 for a population of 1000}
Typically mortality rate in yearly forms of various age classes are given, but a daily or monthly estimate of the rates shall be applied within an agent-based simulation. \\

The desired simulation-adjusted probability is approximated by rather evaluating the desired rate per very short period regardless of the simulation step-size, assumed to be reasonably small (e.g.\ hourly, daily, weekly or monthly at maximum). Formally, the so called instantaneous probability is evaluated as follows:
\begin{equation}
\label{eq:instantaneous}
Pr_{event,instantaneous}(x^t \in X^t,\Delta t) ~=~ - \frac{ln(1-Pr_{event,yearly}(x^t))}{N_{\Delta t}} 
\end{equation}
where $N_{\Delta}$ is given as in Equation \eqref{eq:NDeltaT}.

\section*{Acknowledgments}

The following colleagues are acknowledged 
\begin{itemize}
    \item (Research Associate) Dr.\ Martin Hinsch for scientific exchange
    \item (Research Fellow) Dr.\ Eric Silverman as a principle investigator
\end{itemize}
Both are affiliated at MRC/CSO Social \& Public Health Sciences Unit, School of Health and Wellbeing, University of Glasgow. 
\fi

\section*{Funding}

Dr. Atyiah Elsheikh is a Research Software Engineer at MRC/CSO Social \& Public Health Sciences Unit, School of Health and Wellbeing, University of Glasgow. He is in the Complexity in Health programme. He is supported by the Medical Research Council (MC\_UU\_00022/1) and the Scottish Government Chief Scientist Office (SPHSU16). \\ 

For the purpose of open access, the author(s) has applied a Creative Commons Attribution (CC BY) licence to any Author Accepted Manuscript version arising from this submission.

\bibliographystyle{apalike}
\bibliography{abm}

\end{document}